\documentclass[conference]{IEEEtran}
\IEEEoverridecommandlockouts
\usepackage{cite}
\usepackage{url}
\usepackage{amsmath,amssymb,amsfonts}
\usepackage{algorithmic}
\usepackage{graphicx}
\usepackage{textcomp}
\usepackage{xcolor}
\usepackage{multirow}
\usepackage{subcaption}
\def\BibTeX{{\rm B\kern-.05em{\sc i\kern-.025em b}\kern-.08em
		T\kern-.1667em\lower.7ex\hbox{E}\kern-.125emX}}

\usepackage{pdfpages}
\usepackage{eso-pic}% http://ctan.org/pkg/eso-pic
\definecolor{backcream}{HTML}{ffffef}

\AddToShipoutPictureFG*{% Add picture to background of every page
	\AtPageUpperLeft{%
		\raisebox{-30pt}{\makebox[\paperwidth]{\noindent\fcolorbox{white}{backcream}{\begin{minipage}{18.5cm}\centering\scriptsize
					{\color{red}{This manuscript is accepted to IEEE International Joint Conference on Neural Networks (IJCNN).} {Please cite it as:}}\\\vspace{0.05cm}
					M. Mozafari, L. Reddy and R. VanRullen, ``Reconstructing Natural Scenes from fMRI Patterns using BigBiGAN," 2020 International Joint Conference on Neural Networks (IJCNN), Glasgow, United Kingdom, 2020, pp. 1-8, doi: 10.1109/IJCNN48605.2020.9206960 (\url{https://ieeexplore.ieee.org/document/9206960}).\\\vspace{0.05cm}
					978-1-7281-6926-2/20/\$31.00 \copyright2020 IEEE
				\end{minipage}}}}%
			}
		}

\begin{document}

\title{Reconstructing Natural Scenes from fMRI Patterns using BigBiGAN
\thanks{Funded by AI-REPS ANR-18-CE37-0007-01, ANITI ANR-19-PI3A-0004 and a Nvidia GPU grant.\newline $^{*}$ These authors contributed equally to this work.}
}

\author{\IEEEauthorblockN{Milad Mozafari}
\IEEEauthorblockA{\textit{CerCo, CNRS} \\
%\textit{CNRS}\\
Toulouse, France \\
milad.mozafari@cnrs.fr}
\and
\IEEEauthorblockN{Leila Reddy$^{*}$}
\IEEEauthorblockA{\textit{CerCo, CNRS and} \\
\textit{ANITI, Universit\'e de Toulouse}\\
Toulouse, France \\
leila.reddy@cnrs.fr}
\and
\IEEEauthorblockN{Rufin VanRullen$^{*}$}
\IEEEauthorblockA{\textit{CerCo, CNRS and} \\
\textit{ANITI, Universit\'e de Toulouse}\\
Toulouse, France \\
rufin.vanrullen@cnrs.fr}
% \and
% \and
% \and
}

\maketitle

\begin{abstract}
Decoding and reconstructing images from brain imaging data is a research area of high interest. Recent progress in deep generative neural networks has introduced new opportunities to tackle this problem. Here, we employ a recently proposed large-scale bi-directional generative adversarial network, called BigBiGAN, to decode and reconstruct natural scenes from fMRI patterns. BigBiGAN converts images into a 120-dimensional latent space which encodes class and attribute information together, and can also reconstruct images based on their latent vectors. We computed a linear mapping between fMRI data, acquired over images from 150 different categories of ImageNet, and their corresponding BigBiGAN latent vectors. Then, we applied this mapping to the fMRI activity patterns obtained from 50 new test images from 50 unseen categories in order to retrieve their latent vectors, and reconstruct the corresponding images. Pairwise image decoding from the predicted latent vectors was highly accurate ($\boldmath84\%$). Moreover, qualitative and quantitative assessments revealed that the resulting image reconstructions were visually plausible, successfully captured many attributes of the original images, and had high perceptual similarity with the original content. This method establishes a new state-of-the-art for fMRI-based natural image reconstruction, and can be flexibly updated to take into account any future improvements in generative models of natural scene images.
\end{abstract}

\begin{IEEEkeywords}
fMRI Decoding, Visual Reconstruction, Natural Scenes, BigBiGAN
\end{IEEEkeywords}

\section{Introduction}
For many years, scientists have used machine learning (ML) to decode and understand human brain activity in response to visual stimuli. The great progress of deep neural networks (DNNs) in the last decade has provided researchers with powerful tools and a large number of unexplored opportunities to achieve better brain decoding and visual reconstructions from functional magnetic resonance imaging (fMRI) data.

A variety of approaches have been taken to address image reconstruction from brain data. Before the deep learning era, researchers achieved reconstructions of simple binary stimuli directly from fMRI data~\cite{miyawaki2008visual}. Even though the reconstruction of complex natural images was hardly possible in those days, there were attempts to identify the image within a dataset, instead of reconstructing it: for example, quantitative receptive field models were used to identify the presented image~\cite{kay2008identifying}; in another work~\cite{naselaris2009bayesian}, the authors made use of Bayesian methods to find the image with the highest likelihood.

In recent years, deep networks have brought significant improvements in this field, with the reconstruction of handwritten digits using deep belief networks~\cite{van2010neural}, of face stimuli with variational auto-encoders (VAEs)~\cite{vanrullen2019reconstructing}, and of natural scenes with feed-forward networks~\cite{shen2019deep,beliy2019voxels}, generative adversarial networks (GANs)~\cite{st2018generative, seeliger2018generative}, and dual-VAE/GAN~\cite{ren2019reconstructing}. Most reconstruction methods for natural images, however, tend to emphasize pixel-level similarity with the original images, and rarely produce recognizable objects, or visually plausible or semantically meaningful scenes.

\begin{figure}
	\begin{subfigure}{\columnwidth}
		\centering
		% include first image
		\includegraphics[width=\columnwidth]{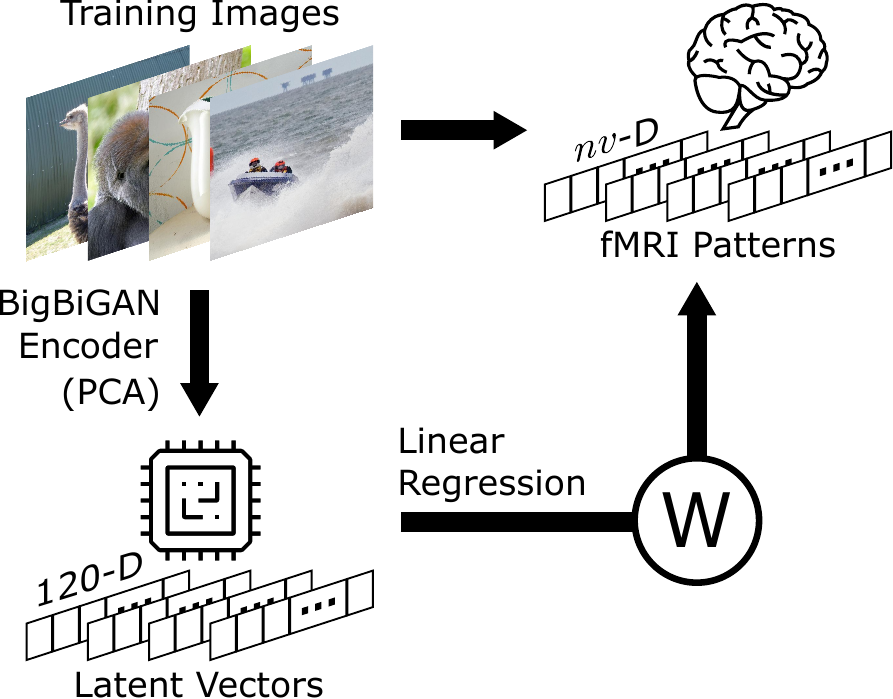}
		\caption{}
		\label{fig:method_a}
	\end{subfigure}
	\begin{subfigure}{\columnwidth}
		\centering
		% include second image
		\includegraphics[width=\columnwidth]{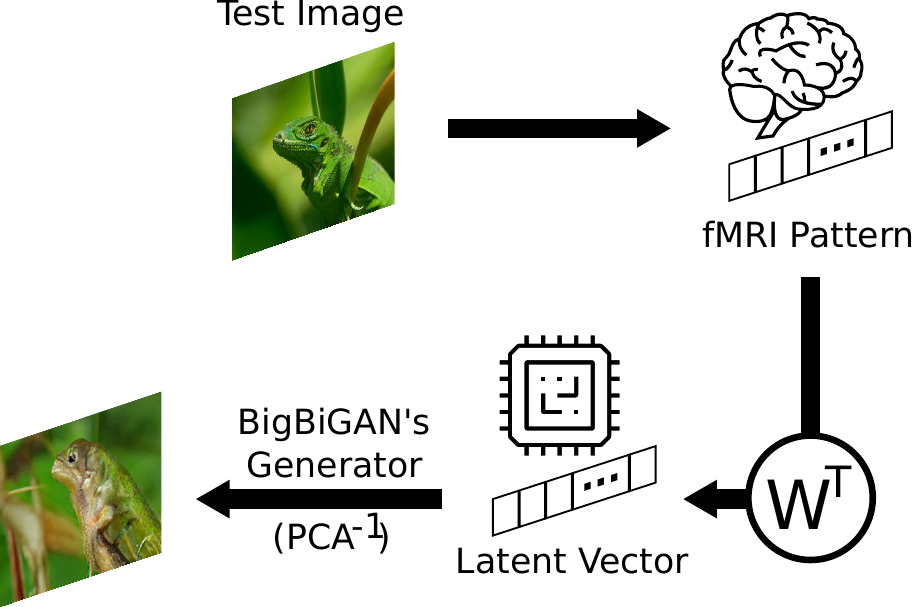} 
		\caption{}
		\label{fig:method_b}
	\end{subfigure}
	\caption{The proposed method. (a) Training phase. We compute a linear mapping (computing the linear transform matrix $W$) from $120$-D latent vectors (derived from the BigBiGAN encoder or from PCA decomposition) to $nv$-D fMRI patterns. $nv$ is the number of voxels inside the brain region of interest. (b) Test phase. The obtained mapping is inversely used to transform fMRI patterns of test images into the latent vectors. The image is then reconstructed using BigBiGAN's generator (or a PCA inverse transform).}
	\label{fig:method}
\end{figure}

Inspired by~\cite{vanrullen2019reconstructing}, we propose a method to reconstruct natural scenes from fMRI data using a recently proposed large-scale bi-directional generative adversarial network, called BigBiGAN~\cite{donahue2019large}. This network is the current state-of-the-art for unconditional image generation on ImageNet in terms of image quality and visual plausibility. In our proposed method, the brain data is mapped to the latent space of the BigBiGAN (pre-trained on ImageNet), whose generator is then used to reconstruct the image. Fig.~\ref{fig:method} demonstrates an overview of the proposed method. Specifically, a training set of natural images that is shown to the human subjects is also fed into BigBiGAN's encoder to get ``original'' latent vectors. Then, a linear mapping is computed between brain responses to the training images and their corresponding original latent vectors. Applying this mapping to the brain data for novel test images, a set of ``predicted'' latent vectors is then generated. Finally, these predicted latent vectors are passed on to the BigBiGAN's generator for image reconstruction.

We demonstrate that the proposed method is able to outperform others by generating high-resolution naturalistic reconstructions thanks to the BigBiGAN generator. We justify our claims by quantitative comparisons of reconstructions to the original images in the high-level representational space of a state-of-the-art deep neural network.

\section{Previous Works}
We begin this section by describing our earlier work from which the method was adapted. In~\cite{vanrullen2019reconstructing}, we took advantage of the latent space of a VAE trained with a GAN procedure on a large set of faces. By learning a linear mapping between fMRI patterns and $1024$-dimensional VAE latent vectors, and using the GAN generator to reconstruct input images, we established a new state-of-the-art for fMRI-based face reconstruction. Moreover, the method even allowed for decoding face gender or face mental imagery.

Despite these promising results on faces, dealing with natural images remains a hard challenge. In another study~\cite{han2018variational}, authors used a VAE for reconstructing naturalistic movie stimuli. They first trained a VAE, with five layers for encoding and five layers for decoding, on the ImageNet dataset. Then, similar to~\cite{vanrullen2019reconstructing}, they converted the fMRI patterns to the VAE's latent space through linear mapping. Although they reported an appreciable level of success, the reconstructions were still blurry and difficult to recognize.

Studies in this field are not limited to the latent space of VAEs. In~\cite{shen2019deep}, the feature space of deep convolutional networks (DCNs) was used for fMRI decoding and image reconstruction. To do so, a decoder was first trained to transform fMRI patterns into the DCN's image representations. Then, for each fMRI pattern, an initial image was proposed and passed through iterative optimization steps. In each iteration, the image was given to the DCN and the difference between its feature representation and the one from the actual image was computed as a loss value. Finally, pixel values were optimized to decrease this loss. The authors also examined optimization in the space of deep generative networks instead of in pixel space. According to the obtained reconstructions, their method was able to capture input attributes such as object color, position, and a coarse estimation of the shape. However, images remained blurry and the objects difficultly recognizable.

Other studies have proposed original network architectures instead of using pre-existing ones. In~\cite{beliy2019voxels} an encoder/decoder structure was proposed, in which the encoder maps images to fMRI data, while the decoder does the reverse. In the first step, the encoder and decoder were separately trained on (image, fMRI) data pairs. Since the number of data pairs was insufficient for proper generalization, the authors applied a second round of training in an unsupervised fashion.

In yet another study~\cite{ren2019reconstructing}, the authors proposed a dual-VAE, trained with a GAN procedure. This method involved three stages of training. In stage 1 the encoder, generator, and discriminator were trained on original images vs. generated ones. In stage 2, the generator was fixed, the  encoder was trained on fMRI data, and the discriminator was trained with reconstructed images from the fMRI data, and reconstructed images from Stage 1. Finally, in Stage 3, the encoder was fixed, and the generator and discriminator networks were fine-tuned using the original images and reconstructed images from the fMRI data. This three-step method not only outperformed previous studies in image decoding, but also generated more crisp and visually plausible reconstructions. However, object identity was not always evident in the reconstructed images.

In this paper, we reconstruct images from human brain activity patterns using the state-of-the-art in natural image generation, a large-scale bi-directional GAN coined ``BigBiGAN''~\cite{donahue2019large}. Notably, the high-level image attributes captured in the latent space of the BigBiGAN allow us to go beyond pixel-wise similarity between the original and reconstructed images, and to reconstruct realistic and visually plausible scenes that express high-level semantic and category-level information from brain activity patterns.

\section{Materials and Methods}

\subsection{fMRI Data}
In this paper, we used open-source fMRI data provided by~\cite{horikawa2017generic}. Images in the stimulus set were selected from ImageNet, and included $1200$ training samples (1 presentation each) from $150$ categories ($150\times 8$), and $50$ test samples (35 presentations each) from $50$ categories. Training and test categories were independent of each other. Five healthy subjects viewed these  training and test images in a fMRI scanner in separate sessions. Each fMRI run consisted of a fixation point ($33$s), $50$ image presentations ($9$s per image, flashing at $2$Hz), and a final fixation point ($6$s). Moreover, $5$ images were randomly repeated during a run and subjects performed a one-back task on these images (i.e., they pressed a button when the same image was presented on two consecutive trials).

We downloaded the raw data\footnote{\url{https://openneuro.org/datasets/ds001246/}} and applied a standard preprocessing pipeline: slice-time correction, realignment, and coregistation to the T$1$w anatomical image using SPM12 software\footnote{\url{https://www.fil.ion.ucl.ac.uk/spm/software/spm12/}}. Details of the parameters used for preprocessing can be found in~\cite{vanrullen2019reconstructing}. The downloaded fMRI dataset also provided pre-defined regions of interest (ROIs) that covered visual cortex. The onset and duration of each image were entered into a general linear model (GLM) as regressors (a separate GLM was used for the training and test sessions).

\subsection{BigBiGAN}
BigBiGAN is a state-of-the-art large-scale bi-directional generative network for natural images~\cite{donahue2019large}. It is a successor of the BiGAN bi-directional GAN~\cite{donahue2016adversarial}, but adopting the generator and discriminator architectures from the more recent BigGAN~\cite{brock2018large}. Similar to BiGAN, the encoder and generator are trained indirectly via a joint discriminator that has to discriminate real from fake [latent vector, data] pairs. The encoder maps data into the latent vectors (real pairs), while the generator reconstructs data from latent vectors (fake pairs). Unlike BigGAN, a conditional GAN which requires a separate ``conditioning'' vector for object category, BigBiGAN's generator has a unified 120-dimensional latent space which captures all properties of objects, including category and pose. In other words, each image can be expressed as a $120$-dimensional vector in the network's latent space, and any latent vector can be mapped back to the corresponding image. The low-dimensionality of the BigBiGAN model makes it particularly appealing for fMRI-based decoding, given the relatively small amount of brain data available for training our system (see~\ref{subsec:decoding}). 

In this study, we used the largest pre-trained BigBiGAN model, revnet50x4, with $256\times 256$ image resolution. The model is publicly available on TensorFlow Hub\footnote{\url{https://tfhub.dev/deepmind/bigbigan-revnet50x4/1}}.

\subsection{PCA Model}
As a baseline image decomposition and reconstruction model for our comparisons, we applied principal component analysis (PCA) on a set of $15000$ images that were randomly selected from the $150$ training categories ($100$ each). We made sure that the $1200$ training images were included. Using the first 120 principal components (PCs), all of the image stimuli were transformed into a set of 120-D vectors. These vectors were then treated similarly to BigBiGAN's latent vectors for brain decoding and reconstructions. This method (known as ``eigen-face'' or ``eigen-image'') has previously been applied to fMRI-based face reconstruction~\cite{cowen2014neural, vanrullen2019reconstructing} and natural image reconstruction~\cite{han2018variational}.

\subsection{Decoding and Reconstruction}\label{subsec:decoding}
Using linear regression, we computed a linear encoder that maps the $120$-dimensional BigBiGAN latent representations (or the $120$-dimensional PCA projections) associated with the training images onto the corresponding brain representations, recorded when the human subjects viewed the same images in the scanner (see Fig.~\ref{fig:method}a). For each subject, this mapping is computed by a general linear regression model (GLM) where the design matrix included the following regressors of interest: fixation (during the fixation point), stimulus (whenever an image was presented), and one-back (when the image was a target for the one-back task). In order to obtain mapping parameters, the $120$-dimensional latent vectors (or PCs) for the training images were added as parametric modulators for the ``stimulus'' regressor. This step takes into account the covariance matrix of the latent dimensions (across images), and produces a linear transform matrix ($W$) which will be used for the inverse transformation in the test phase.

In other words, for the training set of $1200$ images, if there are $nv$ voxels in the desired ROI, the GLM finds an optimal transformation matrix $W$ between their $121$-dimensional latent vectors (including an additional constant bias term) and the corresponding $nv$-dimensional brain activation vectors:
\begin{equation}
Y_{1200\times nv} = X_{1200\times 121} \cdot W_{121\times nv},
\end{equation}
where $X$ and $Y$ denote the latent and brain activation vectors, respectively. Please note that all of the GLMs were solved by SPM12 over the entire visual cortex (union of all pre-defined functional ROIs).

For the test images, brain representations were derived from another GLM in which (in addition to ``fixation'' and ``one-back'' regressors, as previously) the presentation of each test image was considered as a separate regressor. The previously-computed mapping ($W$) was then inverted (again, taking into account the covariance matrix of the latent dimensions, this time across brain voxels), and used to predict the latent vectors (or PCA projections) from the brain representations (see Fig.~\ref{fig:method}b). This corresponds to the ``brain decoding'' step. Precisely, we retrieved the latent vectors $X_{50\times 121}$ from the brain activation vectors of the $50$ test images $Y_{50\times nv}$ using the previously-computed $W$ and its (pseudo-)inverse covariance matrix $(WW^T)^{-1}$:

\begin{equation}\label{eq:test_phase}
    \begin{split}
        Y &= X \cdot W \\
            YW^T &= X \cdot WW^T \\
	    X &= YW^T \cdot (WW^T)^{-1}.
    \end{split}
\end{equation}
Before solving equation~\ref{eq:test_phase}, the brain activation vectors were zero-meaned by subtracting from each the average activation vector across all test images.

Finally, we discarded the bias term from the predicted latent vectors (PCA projections), and fed them into BigBiGAN's generator (PCA's inverse transform) to generate image reconstructions. Since BigBiGAN's generator is sensitive to the distribution of latent variables, we re-scaled predicted latent variables using the mean and standard deviation of latent variables from the training set, before feeding them to the generator. 

\begin{figure*}
    \centering
    \includegraphics[width=0.65\textwidth]{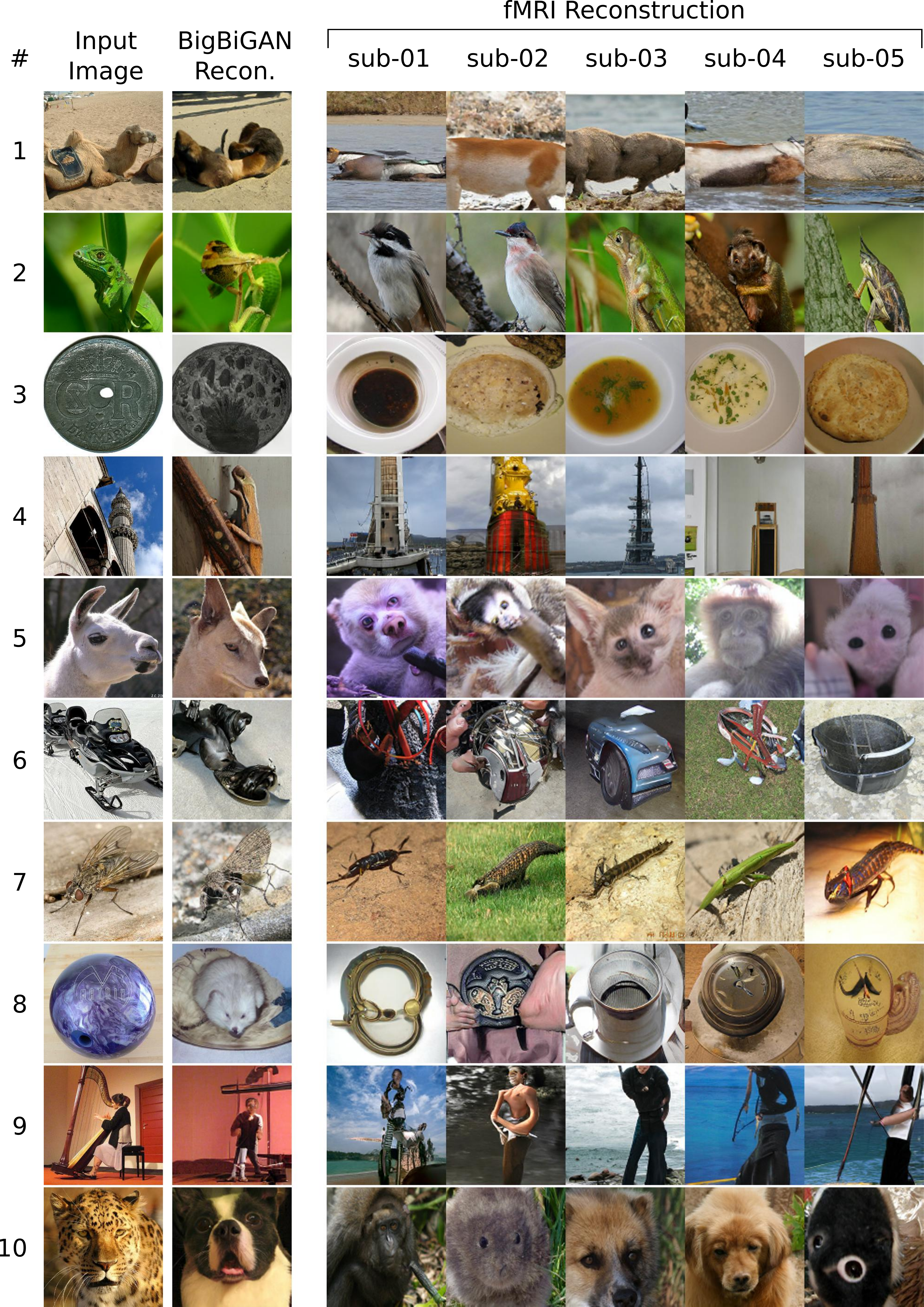}
    \caption{fMRI reconstructions by the proposed method across all subjects. The first and second columns show the input image and BigBiGAN's original reconstruction (reconstruction from the original latent vector), respectively. The next five columns illustrate BigBiGAN's fMRI reconstructions (reconstruction from predicted latent vectors) for each of the five subjects. Although fMRI reconstructions are not a perfect match to the input images, there are many attributes that are consistently captured by all subjects. These attributes can be semantic, such as being an animal or the body pose, and/or visually driven such as roundness or tallness, to mention a few.}
    \label{fig:recon_consistency}
\end{figure*}

\subsection{Computational Efficiency}
The whole computation pipeline from raw fMRI data to image reconstructions consists of the following steps:
\begin{enumerate}
\item fMRI pre-processing
\item Extracting brain representations for test images (GLM)
\item Extracting latent representation for training images (using BiBiGAN's encoder)
\item Computing the linear mapping (GLM)
\item Predicting latent vectors for test images (using the inverse mapping)
\item Reconstructing images (using BiBiGAN's generator)
\end{enumerate}
Apart from the first two steps that are common between almost all fMRI image reconstruction methods, the major computational cost of the proposed method is computing the linear mapping. This is not only considerably less expensive than training large complex encoder/decoder networks (we use pre-trained networks instead), but also easily adaptable to the latent space of any other pre-trained networks. In other words, as soon as a better natural scene generator emerges, we can substitute the new network with the old one and run the pipeline again (from step 2).
For our experiments, we ran this pipeline on a machine running Ubuntu 18.04 with 128 GB of memory, 40 CPU cores (2.20GHz), and NVIDIA TITAN V as the GPU. Nipype python package was also used to parallelize the pre-processing and GLM steps over the five subjects. It took around 16 hours to compute the linear mapping (GLM on the training data) for all subjects, while the encoding and image reconstructions with BigBiGAN took only a few seconds.

\subsection{Decoding Accuracy}
We used a pairwise strategy to evaluate the accuracy of our brain decoder. Assume that there are a set of $n$ (original) vectors $v_1, v_2, ..., v_n$ and their respective predictions $p_1, p_2, ..., p_n$. Then the pairwise decoding accuracy is computed as:
\begin{equation}
\frac{\sum_{i=1}^{n-1}\sum_{j=i+1}^{n} K\left(c(v_i,p_i)+c(v_j,p_j),c(v_i,p_j)+c(v_j,p_i)\right)}{n},
\end{equation}
where $c(.,.)$ is the Pearson correlation and
\begin{equation}
K(a,b) = \begin{cases}
1 & a > b\\
0 & \textit{otherwise}.
\end{cases}
\end{equation}

\begin{figure*}[h]
    \centering
    \includegraphics[width=0.7\textwidth]{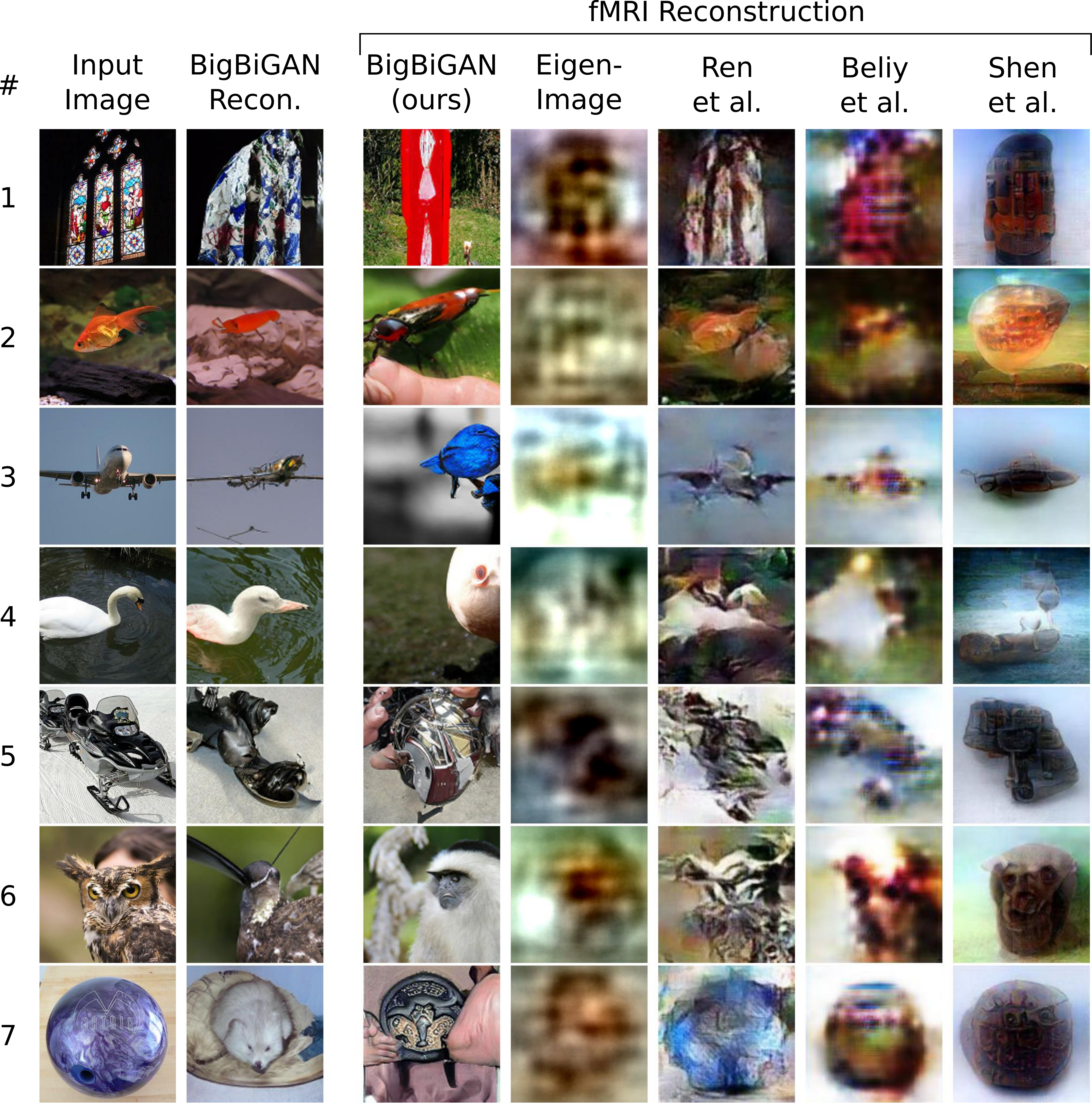}
    \caption{Comparison of fMRI reconstructions by different methods. The first and second columns show the input image and BigBiGAN's original reconstruction (reconstruction from the original latent vector), respectively. Columns three to seven illustrate fMRI reconstructions for BigBiGAN (our method, reconstruction from the predicted latent vector), Eigen-Image (PCA, baseline model), Ren et al.~\cite{ren2019reconstructing}, Beliy et al.~\cite{beliy2019voxels}, and Shen et al.~\cite{shen2019deep}, respectively. Clearly, reconstructions by the proposed method are the most naturalistic, with the highest resolution, in contrast to the more blurry or semantically ambiguous results of the other methods.}
    \label{fig:recon_comp}
\end{figure*}

\subsection{High-Level Similarity Measure}\label{subsec:high-level}
Unlike human judgement, classic similarity metrics such as mean squared error (MSE), pix-comp~\cite{ren2019reconstructing}, or structural similarity index (SSIM) are computed in pixel space and cannot capture high-level perceptual similarities, e.g. in terms of object attributes and identity, or semantic category. One good solution for this problem is to make use of DCN representational spaces, as there are several pieces of evidence supporting their correlation to the human brain~\cite{khaligh2014deep,cichy2016comparison,horikawa2017generic}.

In this paper, Inception-V3~\cite{szegedy2016rethinking} was the DCN of our choice, with the outputs of its last inception block (after concatenation of its branches) defining our high-level representational space. In this space, as a measure of high-level perceptual representations, we computed the average Pearson correlation distance between representations of the original images and their associated fMRI reconstructions. In addition to this high-level measure, we also report pix-comp values~\cite{ren2019reconstructing} as a measure of low-level similarity.

\section{Results}

\subsection{Image Reconstructions}
Using BigBiGAN's generator (or the PCA inverse transform), we could reconstruct an estimate of the test images from the latent vectors obtained by the brain decoder ($W^T$). Since BigBiGAN's generator is not perfect (see first and second columns in Fig.~\ref{fig:recon_consistency} and Fig.~\ref{fig:recon_comp}), we cannot expect the fMRI reconstructions to be identical to the input images (even if our decoding procedure was $100\%$ accurate). However, we found that the brain decoder not only captured several high-level attributes of the images, but that there were robust consistencies in image reconstruction across subjects. Fig.~\ref{fig:recon_consistency} shows a series of reconstructions across all of the five subjects.

For example, when the input image contained an animal (rows 1, 2, 5, 7, 10) or a human (row 9), it was preserved in the reconstructions with comparable location, body shape and pose across subjects. It is worth mentioning that objects or attributes that occur with a higher frequency in the ImageNet dataset are more likely to be preserved in the original BigBiGAN and fMRI reconstructions. For instance, images in the third and eighth rows are not common in Imagenet, yet their roundness attribute is more frequently observed. Thus, all the reconstructions agreed with a round object, even though they could not exactly reconstruct what the object was. Other examples are the images of the tower (fourth row) for the narrowness and tallness attributes, or the insect (seventh row) whose reconstructions mostly captured the long rope-like object behind it and rendered it with insect-related attributes.

\begin{table}[t!]
    \centering
    \caption{Quantitative comparison of image reconstructions. For each measure, the best value is highlighted in bold. (For Pix-Comp, higher is better; for Inception-V3, lower is better)}
    \label{tab:method_cmp}
    \begin{tabular}{|l|c|c|}
        \hline
        \multirow{3}{*}{Method} & \multicolumn{2}{c|}{Similarity Measure}\\
        \cline{2-3}
        ~ & Low-Level & High-Level\\
        ~ & (Pix-Comp) $\uparrow$ & (Inception-V3) $\downarrow$\\
        \hline
        \hline
         Shen et al.~\cite{shen2019deep} & $79.7\%$ & $0.829$ \\
         Beliy et al.~\cite{beliy2019voxels} & $85.3\%$ & $0.865$ \\
         Ren et al.~\cite{ren2019reconstructing} & \boldmath{$87.8\%$} & $0.847$ \\
         Eigen-Image (PCA) & $73.4\%$ & $0.884$ \\
         BigBiGAN (ours) & $54.3\%$ & \boldmath{$0.818$} \\
         \hline
    \end{tabular}
    
\end{table}

fMRI-based natural image reconstruction has been addressed by a variety of methods recently, however only a few of them have been evaluated on the dataset we used. Here, we compare our reconstructions to three recent works by Shen et al.~\cite{shen2019deep}, Beliy et al.~\cite{beliy2019voxels}, and Ren et al.~\cite{ren2019reconstructing}.

Fig.~\ref{fig:recon_comp} shows reconstructions of seven images obtained by each method. Note that we could not compare other images since their reconstructions were not available for all methods. Although our reconstructions are not a perfect match to the input image, they show the clearest resolution, details, and naturalness, and display high-level similarity to the input image. Clearly, PCA (eigen-image) reconstructions rank worst in clarity. The other three methods suffered to varying degrees from ambiguous reconstructions (notably, without any clearly discernible object), although they did much better in estimating low-level attributes of the images, with the best performance obtained by Ren et al. Moreover, unlike the other methods, no image ``halo'' is present in our reconstructions. These halos can result from various factors such as the learning capacity of the encoder/decoder networks, the training approach, and most importantly, pixel-level or low-level similarity optimization, to mention a few.

For a quantitative comparison, we quantified low- and high-level similarities between reconstructions and original images. The former was computed as the pairwise decoding performance in pixel space (pix-comp) for all of the test images, while the latter was the correlation distance between representations of the last inception block in Inception-V3 (see subsection~\ref{subsec:high-level}) over the common set of seven reconstructions showed in Fig.~\ref{fig:recon_comp}. These results (see table~\ref{tab:method_cmp}) justify our claim that high-level aspects of the input images were better preserved by our method, while the other methods had an advantage for low-level aspects.

\subsection{Decoding Accuracy Across Brain Regions}
As mentioned above, the fMRI dataset includes several pre-defined brain regions of interest (ROIs) in visual cortex, including V1 to V4, LOC, FFA, PPA, and HVC as the union of the last three. We also defined the whole visual cortex (VC) as the union of all these ROIs. By limiting voxels to those that were inside each ROI, we evaluated the pairwise decoding accuracy across different regions in visual cortex.

\begin{figure}
    \centering
    \includegraphics[width=\columnwidth]{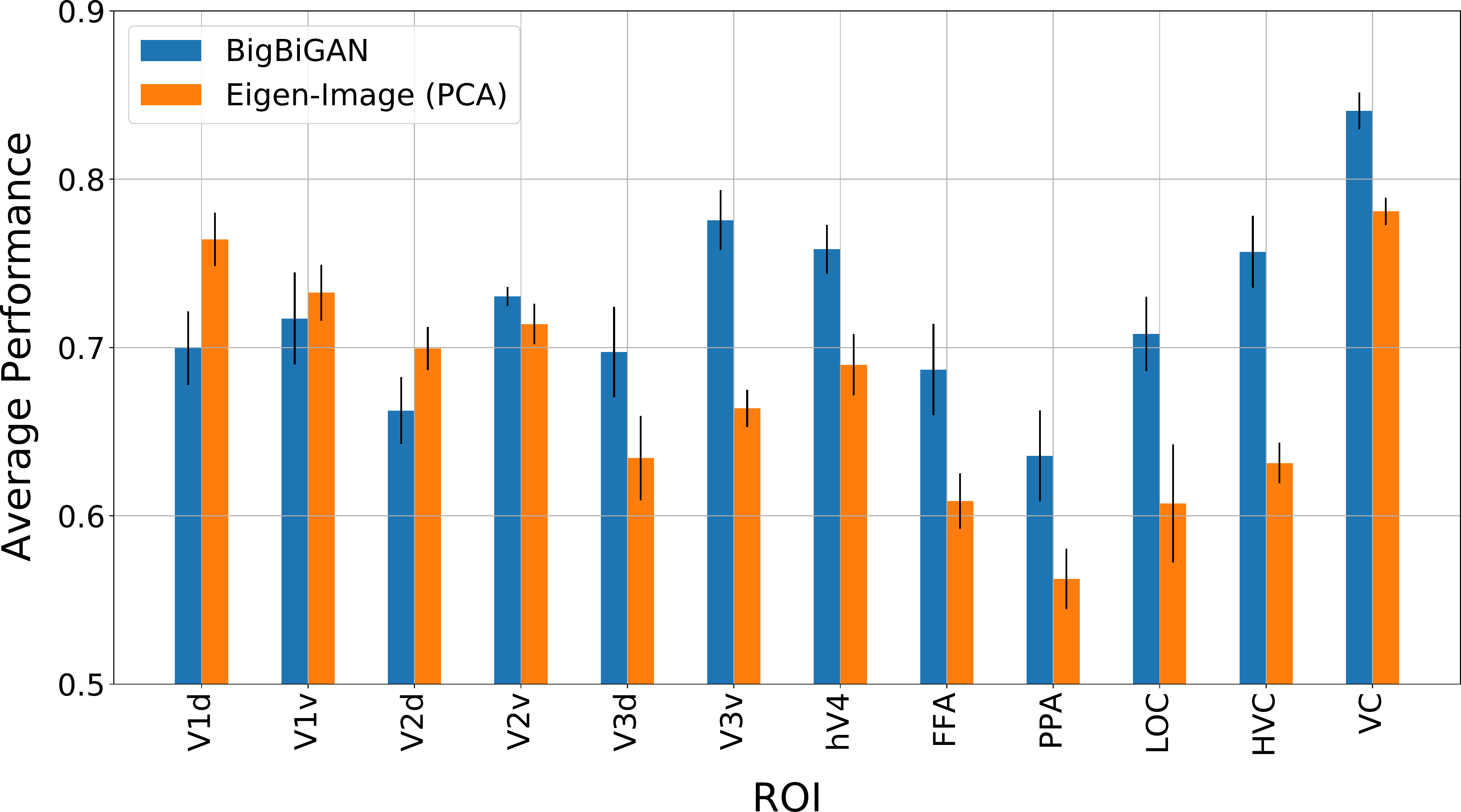}
    \caption{Pairwise decoding accuracy across different brain regions of interest (ROIs). While voxels in high-level areas of the visual cortex are best decoded using BigBiGAN (our method), PCA performs better in low-level regions (V1, V2). Although the best performance is achieved when all the voxels (the whole visual cortex) are included, PCA could only do marginally better than when only V1d voxels were used.}
    \label{fig:decoding_acc}
\end{figure}

Fig.~\ref{fig:decoding_acc} illustrates the average decoding accuracy over all subjects in each brain region. PCA outperformed BigBiGAN in the two earliest visual areas (V1 and V2). However, in higher areas, BigBiGAN gradually improved while PCA worsened.  Peak performance for our method was reached in V3, V4, and HVC, where PCA performed poorly. We hypothesize that the superiority of PCA in lower areas is due to the fact that the PCs were computed in pixel space, and thus correspond mostly to low-level features. On the other hand, BigBiGAN's latent vectors can better represent high-level features, since they are obtained via a large hierarchy of processing layers. For both BigBiGAN and PCA, the best accuracy was achieved when we used brain responses from the whole VC. Peak accuracy was $84.1\%$ and $78.1\%$ for BigBiGAN and PCA, respectively.

It is worth mentioning that, while the whole VC improved BigBiGAN's performance significantly compared to each individual region, PCA could only do marginally better than when using voxels in V1d alone (its best single-region performance). This again suggests that PCA mostly depends on low-level features, whereas the BigBiGAN brain decoder can benefit from low-level information as well as high-level image attributes.

\section{Discussion}
In this paper, we have proposed a new method for realistic reconstruction of natural scenes from fMRI patterns. Thanks to the high-level, low-dimensional latent space of BigBiGAN, we could establish a linear mapping that associates image latent vectors to their corresponding fMRI patterns. This linear mapping was then inverted to transform novel fMRI patterns into BigBiGAN latent vectors. Finally, by feeding the obtained latent vectors into the BigBiGAN generator, the associated images were reconstructed.

Many recent approaches have taken advantage of deep generative neural networks to reconstruct natural scenes~\cite{shen2019deep, beliy2019voxels, ren2019reconstructing}. However, due to the complexity of natural images, a huge amount of computational resources and capacity is required to achieve high-resolution realistic image generation~\cite{brock2018large}. Here, we used the pre-trained BigBiGAN as a state-of-the-art large-scale bi-directional GAN for natural images. We showed that the proposed method is able to generate the most realistic reconstructions in the highest resolution ($256\times 256$) compared to other methods. Moreover, comparing results across subjects revealed a robust consistency in capturing high-level attributes of different objects through the reconstructions.

We acknowledge that our reconstructions are still far from perfect and can often lag behind the others in terms of low-level similarity measures. In contrast, the superiority of the proposed method is with respect to high-level evaluations of perceptual similarity. While we can surpass other methods in this area, we believe that there is still room for methodological improvements. In particular, failures to retrieve the proper semantic category or visual attribute can of course be caused by imperfect brain-decoding of the latent vectors, but also sometimes by inadequate image generation from the BigBiGAN generator (e.g., compare the first 2 columns in Fig.~\ref{fig:recon_consistency}). We believe that one promising area of improvement for our work is through the ability of the image generation model. In this regard, whenever new bidirectional GANs (or other bidirectional architectures) improve on the current state-of-the-art, our method can easily be adapted to deploy them and take advantage of their image generation prowess for more accurate brain-based reconstructions.

Another current limitation of the proposed method is our use of pre-defined brain regions of interest (or potentially, of the entire visual cortex). It is likely that not all voxels are informative or relevant to the target task; including uninformative or irrelevant voxels can only degrade the outcome. Additionally, there might well be informative voxels in other brain areas such as pre-frontal cortex, signaling high-level perceptual or semantic aspects of the visual stimulus, that we are currently not considering. For these reasons, extending the analysis to the entire brain, while using a proper voxel selection stage to discard irrelevant voxels, is bound to further improve the results.

%\section*{Acknowledgment}

% \bibliographystyle{IEEEtran}
% \bibliography{ref_rev1}

\end{document}